\documentclass[twocolumn,10pt,a4paper]{article}

\usepackage{graphicx}
\usepackage{CJKutf8} 
\usepackage{tikz} 
\usepackage{amsmath}
\usepackage{multirow}
\usepackage{hyperref}
\usepackage{url}
\usepackage{authblk}
\usepackage[top=25mm,bottom=30mm,left=25mm,right=25mm]{geometry}

\makeatletter
\renewcommand{\@makefntext}[1]{%
    \noindent
    \makebox[0pt][r]{\@thefnmark\ }#1

    \addvspace{6pt}
}
\makeatother

\title{Investigating the structure of emotions by analyzing similarity and association of emotion words}

\author[]{Fumitaka Iwaki}
\author[]{Tatsuji Takahashi}
\affil[]{School of Science and Technology, Tokyo Denki University}
\date{}

\begin{document}

\renewcommand{\thefootnote}{}  

\maketitle

\footnotetext[1]{%
    F. Iwaki \\
    Ishizaka, Hatoyama, Hiki, Saitama, 350-0394, JAPAN \\
    E-mail: 25udj01@ms.dendai.ac.jp
}
\footnotetext[2]{%
    T. Takahashi \\
    E-mail: tatsuji.takahashi@gmail.com
}
\footnotetext[3]{%
    The dataset created and used in this study is publicly available at \url{https://github.com/FumitakaIwaki/emotion_network_data.git} .
}
\renewcommand{\thefootnote}{\arabic{footnote}}  
\setcounter{footnote}{0}

\begin{abstract}
In the field of natural language processing, some studies have attempted sentiment analysis on text by handling emotions as explanatory or response variables. 
One of the most popular emotion models used in this context is the wheel of emotion proposed by Plutchik. 
This model schematizes human emotions in a circular structure, and represents them in two or three dimensions. 
However, the validity of Plutchik's wheel of emotion has not been sufficiently examined. 
This study investigated the validity of the wheel by creating and analyzing a semantic networks of emotion words. 
Through our experiments, we collected data of similarity and association of ordered pairs of emotion words, and constructed networks using these data.
We then analyzed the structure of the networks through community detection, and compared it with that of the wheel of emotion. 
The results showed that each network's structure was, for the most part, similar to that of the wheel of emotion, but locally different.


\end{abstract}

\section{Introduction}
\label{sec:intro}
A network is a useful representation for expressing the structure of relationship such as similarity and association between entities.
Inohara and Utsumi studied the similarity and association of commonly Japanese word pairs and created a Japanese dataset.
They indicated that while there is a high correlation between similarity and association, there are differences in their properties \cite{inohara2022}.

In the fields of natural language processing and \newline robotics, studies concerning human emotions are abundant \cite{acheampong2020}.
In treating emotions as explanatory or response variables, some of these studies adopted Plutchik's wheel of emotion as the structure of emotions \cite{plutchik2001}. 
For example, Kumar and Vardhan have conducted sentiment analysis of text based on the wheel of emotion \cite{kumar2022}. 
Qi et al. proposed an emotional model that includes emotions, moods, and characters based on the wheel of emotion, and created a social robot that imitates emotional changes of humans \cite{qi2019}.
However, there is more room for the validity of the structure of the wheel of emotion.
Yamashita and Kudoh examined the validity of the wheel model using ElectroEncephaloGraphy and facial expression \cite{yamashita2022}.
In similar ways, some studies have attempted to extract the characteristics of each emotion and then used them to construct the overall structure.
In contrast, we thought that the method that extracts features from links between emotions might be more effective.
However, there have been no attempts to examine the validity of the structure from the viewpoint of relationships between emotions. 

In this study, we have attempted to examine the validity of the emotion model by constructing and analyzing semantic networks of emotions.
Firstly, we collected data on the similarity and association in terms of evaluation of the ordered pairs of emotion words.
We constructed the networks from the data on the similarity and association between the ordered pairs of emotion words obtained through our experiments. 
We then pointed out the differences in the nature of the similarity and association by comparing the locality and globality of those two types of datasets.
Then, we decomposed the networks to communities to analyze the structure. 
We used modular decomposition of Markov chain (MDMC), which enabled us to extract pervasive and hierarchical communities in various networks, as a method of community detection \cite{okamoto2022}.
MDMC assumes that each node probabilistically belongs to all communities and provides the probability distribution of communities for each node.
We can obtain, in the other way around, the probability distribution of nodes for each community through this method, too. 
It also has a parameter of resolution called $\alpha$ for adjusting the size and number of communities.
Finally, we examined the structural validity of the model from comparison with the two types of structures.

In recent years, research on Human-Agent Interaction (HAI) has gained momentum.
We think that clarifying the relationships between emotions is an important part to create more natural and human-friendly social agents.
This paper is a fundamental research for achieving HAI technologies that can include emotions.

\section{The wheel of Emotion}
\label{sec:wheel-of-emotion}
\begin{figure}[htbp]
    \includegraphics[width=\columnwidth]{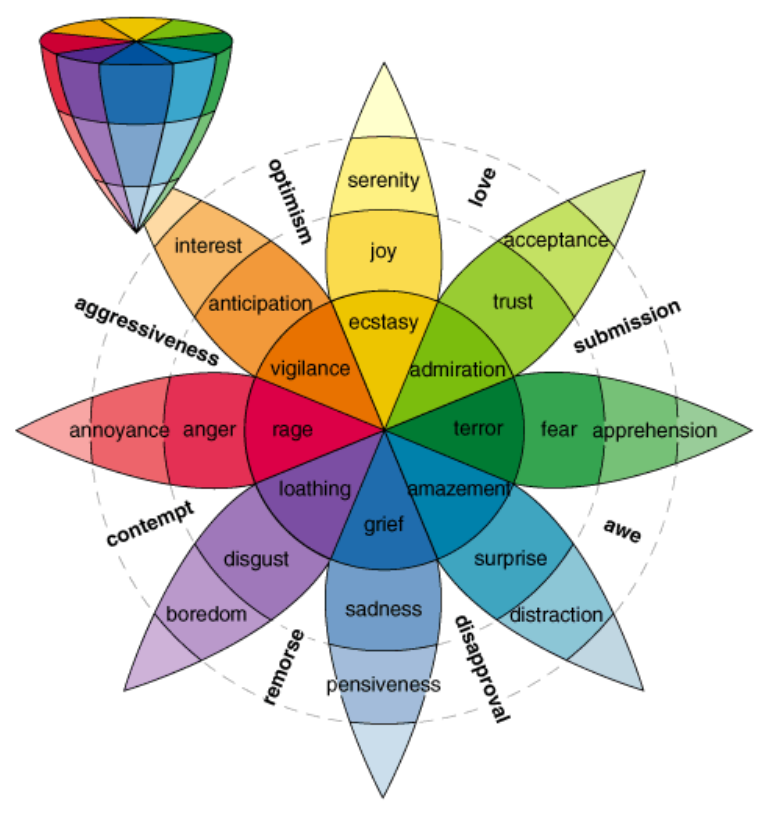}
    \caption{The wheel of emotion © Robert Plutchik / \newline\protect\url{https://www.fractal.org/Bewustzijns-Besturings-Model/Nature-of-emotions.htm} / CC-BY-SA-3.0}
    \label{fig:plutchik_wheel}
\end{figure}

One of the most popular emotion models is the wheel of emotion proposed by Plutchik \cite{plutchik2001}. 
This model schematizes human emotions in a circular structure, and allocates eight primary emotions on it. 
The inner side of the circle has strongly expressed emotions of those primary emotions, and the outer side of the circle has weakly expressed emotions. 
Secondary emotions made by mixing two adjacent primary emotions are located between them.  
A pair of emotions next to each other are similar, and an emotion and another across from it are opposite emotions.
Fig. \ref{fig:plutchik_wheel} shows the wheel of emotion. 
Plutchik also suggested that there are secondary emotions derived from a pair of primary emotions that are not adjacent to each other.
In this study, we used these 48 emotion words proposed by Plutchik for our experiments.

\vspace{10mm}
\begin{table}[htbp]
    \centering
    \caption{48 emotion words proposed by Plutchik}
    \resizebox{\columnwidth}{!}{%
    \begin{tabular}{ll}
        \hline\noalign{\smallskip}
        \multirow{3}{*}{primary emotions} & joy trust fear surprise \\
        & sadness disgust anger \\
        & anticipation \\
        \noalign{\smallskip}\hline
        \multirow{3}{*}{strong derived emotions} & ecstasy admiration terror \\ &amazement grief loathing \\
        & rage vigilance \\
        \noalign{\smallskip}\hline
        \multirow{3}{*}{weak derived emotions} & serenity acceptance apprehension \\
        & distraction pensiveness boredom \\
        & annoyance interest \\
        \noalign{\smallskip}\hline
        \multirow{7}{*}{secondary emotions} & optimism hope anxiety love guilt \\
        & delight submission curiosity \\
        & sentimentality awe despair shame \\
        & disappointment unbelief outrage \\
        & remorse envy pessimism \\
        & contempt cynicism morbidness \\
        & aggressiveness pride dominance \\
        \noalign{\smallskip}\hline
    \end{tabular}%
    }
    \label{tab:emotions}
\end{table}

\section{MDMC}
\label{sec:mdmc}
Modular decomposition of Markov chain (MDMC) is one of the methods for community detection. 
This method involves considering a random walk on a network and identifying communities based on where a random walk tends to linger.
Let us assume that node $i$ has a probability $p(i)$ and community $k$ has a probability $\pi(k)$.
And, $p(i|k)$ denote a conditional probability of node $i$ given community $k$.
Then, we can assume $p(i)$ is dissolved by $\pi(k)$ and $p(i|k)$ as follows:(\ref{fom:p(i)}).
\begin{equation}
    \label{fom:p(i)}
    p(i) = \sum_{k=1}^K \pi(k)p(i|k)
\end{equation}
%
We can decompose networks into communities by solving for $\pi(k)$ and $p(i|k)$ using $p(i)$ as input. 
In order to estimate the value of $\pi(k)$ and $p(i|k)$, we introduce a posterior probability and maximize it using expectation–maximization (EM) algorithm. 
The algorithm has the parameter $\alpha$ in its maximization step.  
This represents the resolution parameter to adjust the number and size of communities. 
The smaller values $\alpha$ are, the decomposed network results in a greater number of and smaller sizes of communities.
Lastly, $p(k|i)$ is defined using Bayes' theorem as follows:(\ref{fom:p(k|i)}).
\begin{equation}
    \label{fom:p(k|i)}
    p(k|i) = \frac{p(i|k)\pi(k)}{p(i)}
\end{equation}
%
Generally, $p(k|i)$ is a positive number for multiple $k$. 
This indicates the pervasiveness of a node $i$, meaning that it belongs to multiple communities. 
When we uniquely define the community to which node $i$ belongs, we obtain it by argmax$_kp(k|i)$.

\section{Experiments}
\label{sec:experiments}
We conducted two experiments in the similar way with the research on the similarity and association of Japanese word pairs and the study on the resemblance of the colors \cite{inohara2022,kawakita2023}. 
We have conducted two experiments independently, one which examined similarity of the ordered pairs of the emotion words and the other which examined association of it. 
At the beginning of each experiment, we explicitly described the participants the difference between similarity and association.
Then, we made networks based on the data obtained by these experiments.
The dataset is publicly available at \url{https://github.com/FumitakaIwaki/emotion_network_data.git} .

\subsection{Material and method}
\label{sec:experiment-method}
We conducted the two experiments online using a web application. 
For the development of the web application, we used Flask, which is a web application framework of Python.
We hosted the server for the experiments on Amazon Web Service EC2. 
We recruited participants through a crowdsourcing platform called CrowdWorks (\url{https://crowdworks.jp/}).
%

We used 48 emotion words in Table \ref{tab:emotions}. 
For each ordered pair of these words, we asked participants about the degree of similarity and association.
The procedures for the similarity experiment was as follows: we presented participants with the questions ``How similar is A to B?" for all two emotions A and B, and they then dragged a slider to choose an evaluation score from 0 (not at all) to 7 (very similar) . 
In the experiment of association, we presented questions ``How associated is A with B?" and participants chose an answer from the same evaluation scores.
To remove poor quality answers from the obtained data, we applied two kinds of filtering, catch trials and a double-pass procedure.
In catch trials, we inserted the request ``Please choose $n$" twice during each experiment ($n$ was a value from 0 to 7) .
We removed the responses of the participants who have not properly chosen the indicated value at least once of the catch trials. 
In the double-pass process, we again subjected to the participants 20 questions from within the questions that have already been submitted at the end of the experiments. 
We deleted the data of which the correlation coefficient of the 20 pairs of answers was less than 0.4.

The participants for each experiment were 360 people, whose ages ranged from 20s to 70s. 
%
First, we removed defective data from the data we collected.
We deleted 27.22\% of the similarity data and 15.56\% of the association data, respectively.
We furthermore applied the aforementioned two types of filtering to these data.
We removed 2.29\% of the data in catch trials and 6.87\% of the data in double-pass from similarity data.
From association data, we removed 2.63\% of the data in catch trials and 24.01\% of the data in double-pass.
%
Finally, we used the data of 240 people (age:$M = 42.5$, $SD = 8.9$, sex:male 131, female 108, others 1) from the experiment of similarity and the data of 229 people (age:$M = 43.5$, $SD = 10.1$, sex:male 116, female 112, others 1) from the experiment of association.
\subsection{Construction of networks}
\label{sec:const-network}
We constructed two networks based on the data obtained by the experiments.
The nodes of both networks correspond to 48 emotion words.
We set its edge weights to the average of evaluation scores of similarity or association.

\section{Result}
\label{sec:result}

\subsection{Differences between similarity and association}
\label{sec:differences-sim-asc}
We compared the distributions of the evaluation scores between similarity and association to express clearly the differences in properties of them. 
We found that there were more responses with a evaluation score of 0 in similarity compared to association.
On the other hand, the higher the evaluation value, the more the number of responses of association than similarity we obtained. 
We tested the difference in the distribution of responses of similarity and association by chi-square test, and we recognized significantly different at a 5$\%$ significance level ($\chi^2$-statistic $= 1091.37$, $p = 2.16\times10^{-231}$, power $= 1.0$, df $= 7$) .

We defined locality of the networks as the strength of connection between words in same petal on the wheel of emotion. 
The petals are the color-coded ellipses that constitute the wheel model. 
We defined globality of the networks as the strength of relations between a pair of petals directly opposite across from each other. 
Then, we compared similarity and association on locality and globality. 
$L_{w_1, w_2}$ represents the weight of the link from word $w_1$ to word $w_2$. 
$W_k$ represents the set of the words within petal $k$. 
$\overline{W}_k$ represents the set of the words within the petal opposite to petal k. 
We defined the set of the pairs of the words jas follows: (\ref{fom:S_k}) and (\ref{fom:S_k_bar}).
\begin{eqnarray}
    \label{fom:S_k}
    S_k &=& \{(w_1, w_2)|w_1, w_2 \in W_k, w_1 \neq w_2\} \\
    \label{fom:S_k_bar}
    \overline{S}_k &=& \{(w_1, w_2)|w_1 \in W_k, w_2 \in \overline{W}_k\}
\end{eqnarray}
%
The maximum possible weight $E_{max}$ was 7, and the number of the petals $K$ was 8. 
We formulated locality and globality of the networks by using the above factors in the following.
\begin{align}
    \label{fom:locality}
    \mathrm{locality} &=& \frac{1}{K}\sum_{k=1}^K \frac{1}{|S_k|}\sum_{w_1, w_2 \in S_k} \frac{L_{w_1, w_2} + L_{w_2, w_1}}{2E_{max}} \\
    \label{fom:globality}
    \mathrm{globality} &=& \frac{1}{K}\sum_{k=1}^K \frac{1}{|\overline{S}_k|}\sum_{w_1, w_2 \in \overline{S}_k} \frac{L_{w_1, w_2} + L_{w_2, w_1}}{2E_{max}}
\end{align}
%
(\ref{fom:locality}) takes 1 when the edge weights are $E_{max}$ in all pairs $S_k$. (\ref{fom:globality}) also takes 1 the edge weights are $E_{max}$ in all pairs $\overline{S}_k$. 
Locality represents how close the edge weights of all word pairs in same petal are to $E_{max}$. Globality represents how close the edge weights of all word pairs between the petals facing each other are to $E_{max}$.
We show the calculation results of locality and globality produced from the above formulas in Table \ref{tab:local_global}.
\begin{table}[htbp]
    \caption{Values of locality and globality}
    \begin{tabular}{lrr}
        \hline\noalign{\smallskip}
         & locality & globality \\
         \hline\noalign{\smallskip}
         similarity & .60 & .41 \\
         association & .68 & .53 \\
         \hline\noalign{\smallskip}
    \end{tabular}
    \label{tab:local_global}
\end{table}
%
It shows that both locality and globality were higher for association, with a more pronounced difference observed in globality.
We measured the differences of similarity and association in regards to locality and globality with $t$-test.
%
We acknowledged significant differences at a $5\%$ significance level in both of them (locality: $t$-statistic $= -4.60$, $p = 3.21\times10^{-5}$, power $= 0.49$, df $= 47$, globality: $t$-statistic $= -5.94$, $p = 3.32\times 10^{-14}$, power $= 0.93$, df $= 71$).

\subsection{Decomposition of the networks}
\label{sec:decompose-network}
\begin{table}[bp]
    \caption{The state of communities in the similarity network (8 communities)}
    \begin{tabular}{ll}
        \hline\noalign{\smallskip}
         k & member \\ 
         \hline\noalign{\smallskip}
         \multirow{2}{*}{1} & joy surprise ecstasy amazement admiration interest\\ 
         & delight curiosity awe pride \\
         \hline\noalign{\smallskip}
         \multirow{2}{*}{2} & trust anticipation serenity acceptance optimism hope \\
         & love aggressiveness dominance \\
         \hline\noalign{\smallskip}
         \multirow{2}{*}{3} & fear terror apprehension distraction vigilance anxiety \\
         & unbelief \\
         \hline\noalign{\smallskip}
         \multirow{2}{*}{4} & sadness grief pensiveness guilt sentimentality despair \\
         & shame disappointment remorse pessimism morbidness \\
         \hline\noalign{\smallskip}
         \multirow{2}{*}{5} & disgust anger loathing rage annoyance outrage envy \\
         & contempt \\
         \hline\noalign{\smallskip}
         6 & boredom \\
         \hline\noalign{\smallskip}
         7 & submission \\
         \hline\noalign{\smallskip}
         8 & cynicism \\
         \hline\noalign{\smallskip}
    \end{tabular}
    \label{tab:sim_comm_8}
\end{table}
\begin{table}[bp]
    \caption{The state of communities in the association network (8 communities)}
    \begin{tabular}{ll}
        \hline\noalign{\smallskip}
         k & member \\ 
         \hline\noalign{\smallskip}
         \multirow{3}{*}{1} & joy surprise anticipation ecstasy admiration amazement \\
         & interest delight curiosity awe aggressiveness \\ 
         & pride dominance \\
         \hline\noalign{\smallskip}
         2 & trust serenity acceptance optimism hope love\\
         \hline\noalign{\smallskip}
         \multirow{2}{*}{3} & fear terror apprehension grief vigilance anxiety guilt \\
         & submission despair unbelief morbidness \\
         \hline\noalign{\smallskip}
         4 & sadness distraction pessimism \\
         \hline\noalign{\smallskip}
         \multirow{2}{*}{5} & disgust anger loathing rage annoyance disappointment \\
         & outrage envy contempt \\
         \hline\noalign{\smallskip}
         6 & pensiveness sentimentality \\
         \hline\noalign{\smallskip}
         7 & boredom cynicism \\
         \hline\noalign{\smallskip}
         8 & shame remorse \\
         \hline\noalign{\smallskip}
    \end{tabular}
    \label{tab:asc_comm_8}
\end{table}
We performed community decomposition using MDMC to the similarity and association networks made from the data. 
We predetermined the total number of communities $K$ to be 10, since MDMC needs it to be fixed in advance. 
When we set the parameter of resolution $\alpha$ as .001, both networks were decomposed to 8 communities.
We show the result of the detection in Table \ref{tab:sim_comm_8} and \ref{tab:asc_comm_8}. 
$k$ represents the index for communities, allocated arbitrarily.

\begin{figure}[htbp]
    \includegraphics[width=80mm]{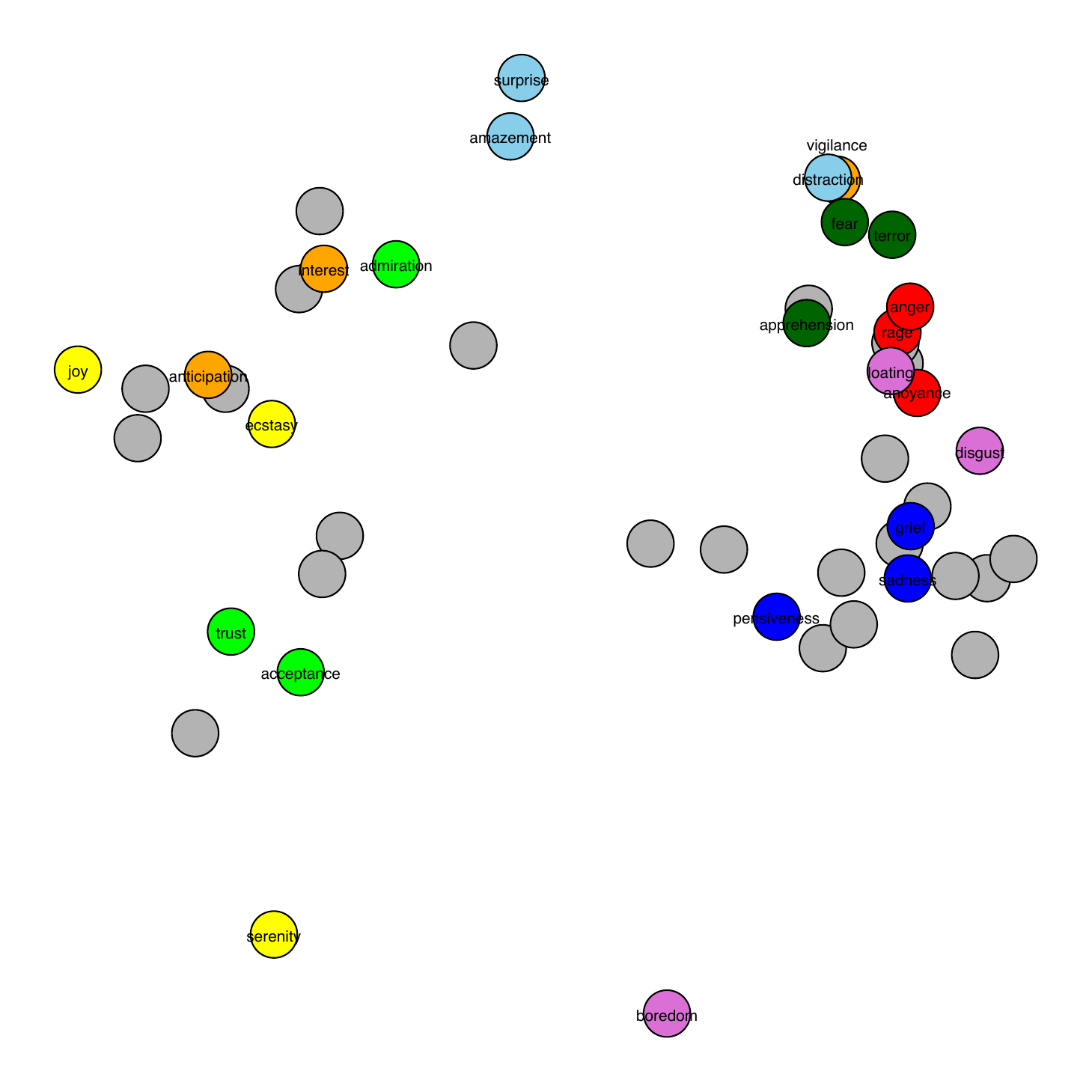}
    \caption{The similarity network (8 communities)}
    \label{fig:sim_network_8}
\end{figure}

\begin{figure}[htbp]
    \includegraphics[width=80mm]{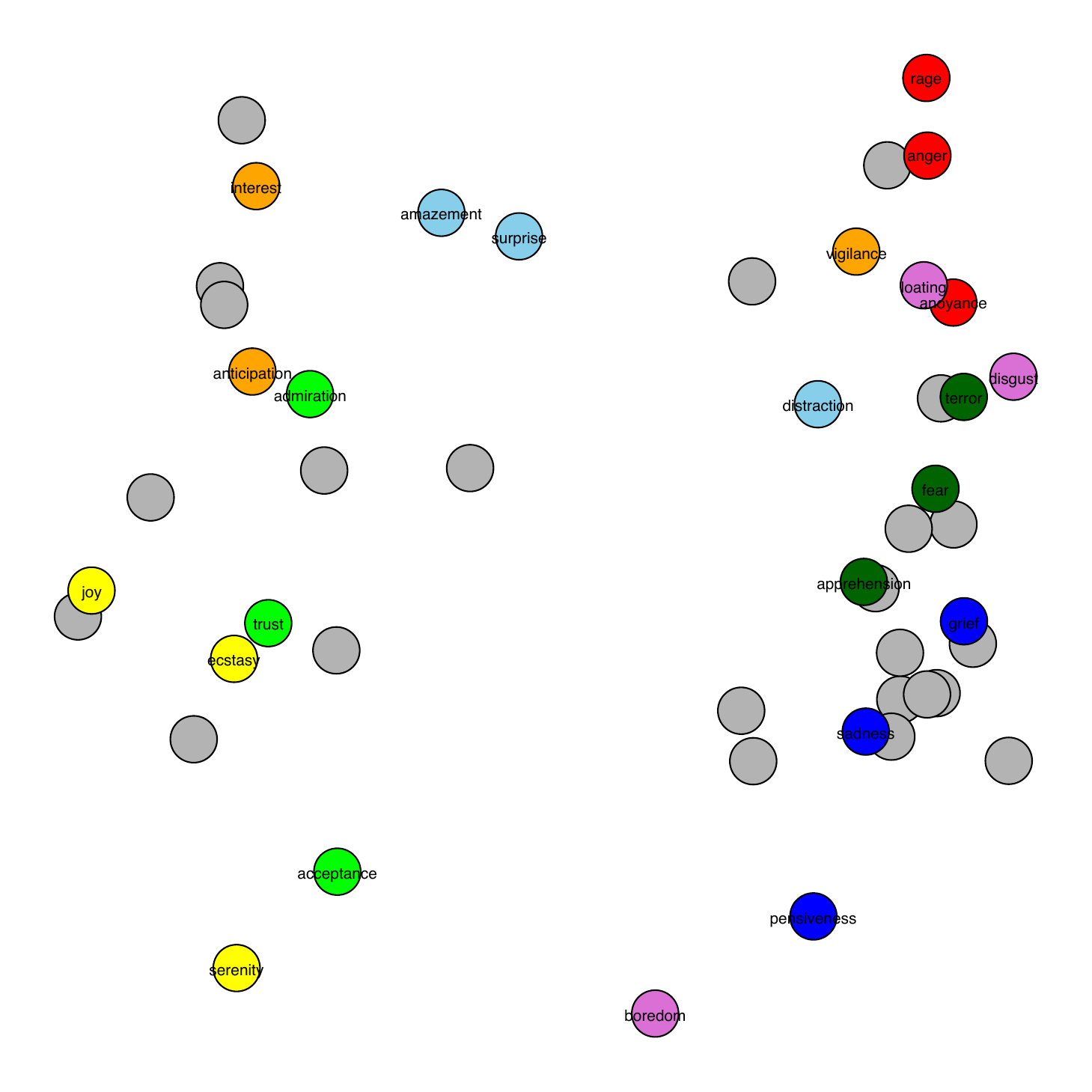}
    \caption{The association network (8 communities)}
    \label{fig:asc_network_8}
\end{figure}

We made two dissimilarity matrices from the scores of similarity or association to plot the networks with multidimensional scaling (MDS). 
Then, we colored the nodes that correspond to the primary emotions and the strong and weak derived emotions with the matching colors in the wheel of emotion. 
We illustrate the product of it in Fig. \ref{fig:sim_network_8} and \ref{fig:asc_network_8}.
%
Using normalized mutual information (NMI), we compared these results of community detection with the structure of the wheel of emotion. 
NMI is a indicator that represents the difference between two clusters. 
If two clusters have mutual information, NMI takes a value close to 1; otherwise, if they do not, NMI value approaches 0. 
In the wheel of emotion, we regarded each petal as a separate community.
As the result of this comparison, the NMI that the wheel of emotion to similarity network was 0.81. 
Likewise, in association network it was 0.72. 
It turned out that the similarity network was slightly more similar to the wheel model than the association network.

\subsection{The relationship between communities}
\label{sec:relationship-communities}
We quantified the strength of the connections between the communities and made the networks whose nodes are community. 
Okamoto formulated the strength of a connection between two communities \cite{okamoto-p-c}. The strength of the connection from community $k$ to community $k'$ is given as (\ref{fom:Omega}).
\begin{eqnarray}
    \label{fom:Omega}
    \Omega_{k'k} = \sum_{i=1}^N\sum_{j=1}^N \pi(k')p(i|k') T_{ij} \pi(k)p(j|k)
\end{eqnarray}
Where $\pi(k)$ is the probability of $k$ that is calculated in the process of MDMC, $T_{ij}$ is the probability of transition from node $j$ to node $i$ that is calculated using PageRank algorithms, and $p(i|k)$ is the conditional probability of the node $i$ given community $k$.
$N$ represents the number of nodes.
$\Omega$ is an asymmetric index ($\Omega_{k'k} \neq \Omega_{kk'}$).

\begin{figure}[htbp]
    \includegraphics[width=80mm]{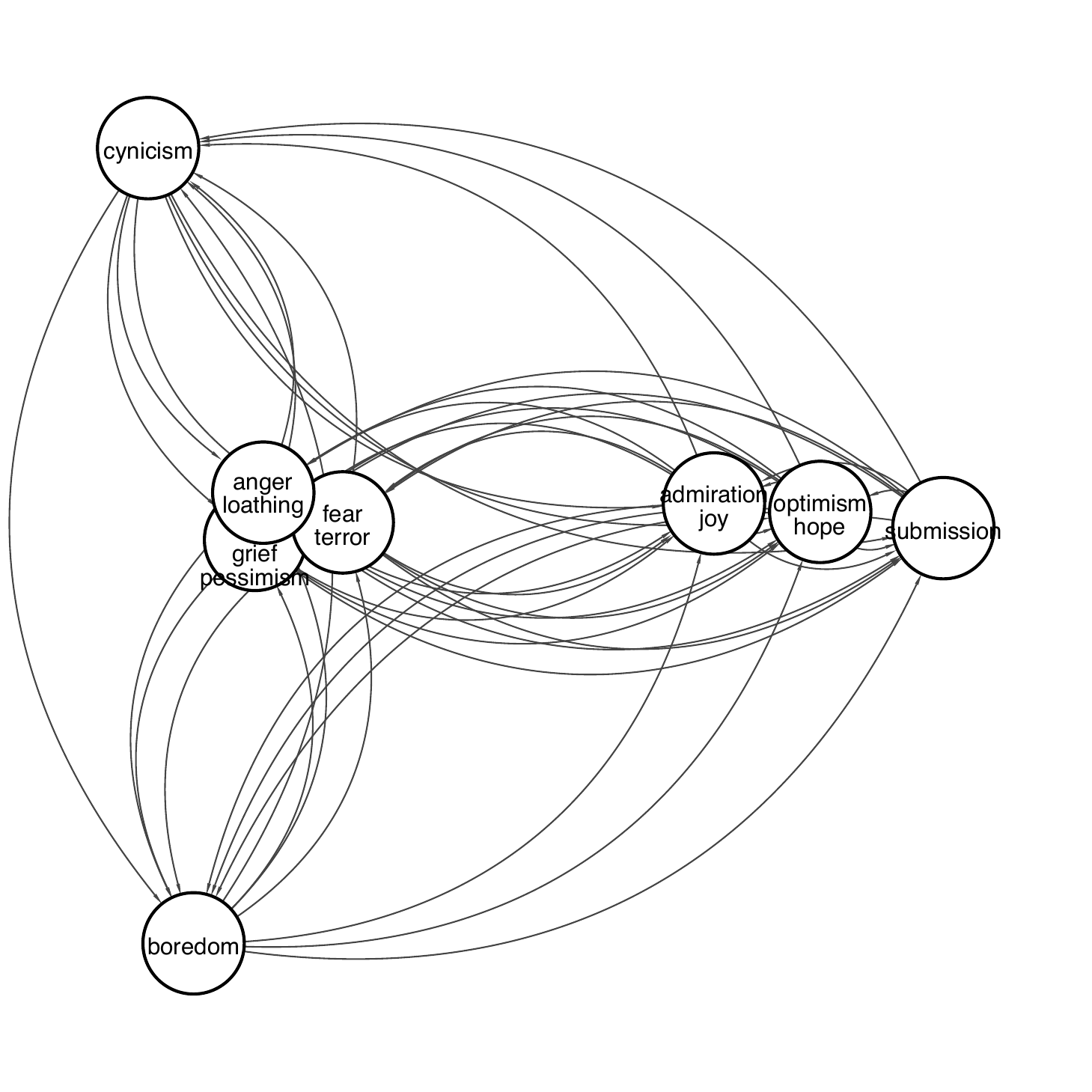}
    \caption{The network of community based on the similarity}
    \label{fig:sim_communities_network}
\end{figure}

\begin{figure}[htbp]
    \includegraphics[width=80mm]{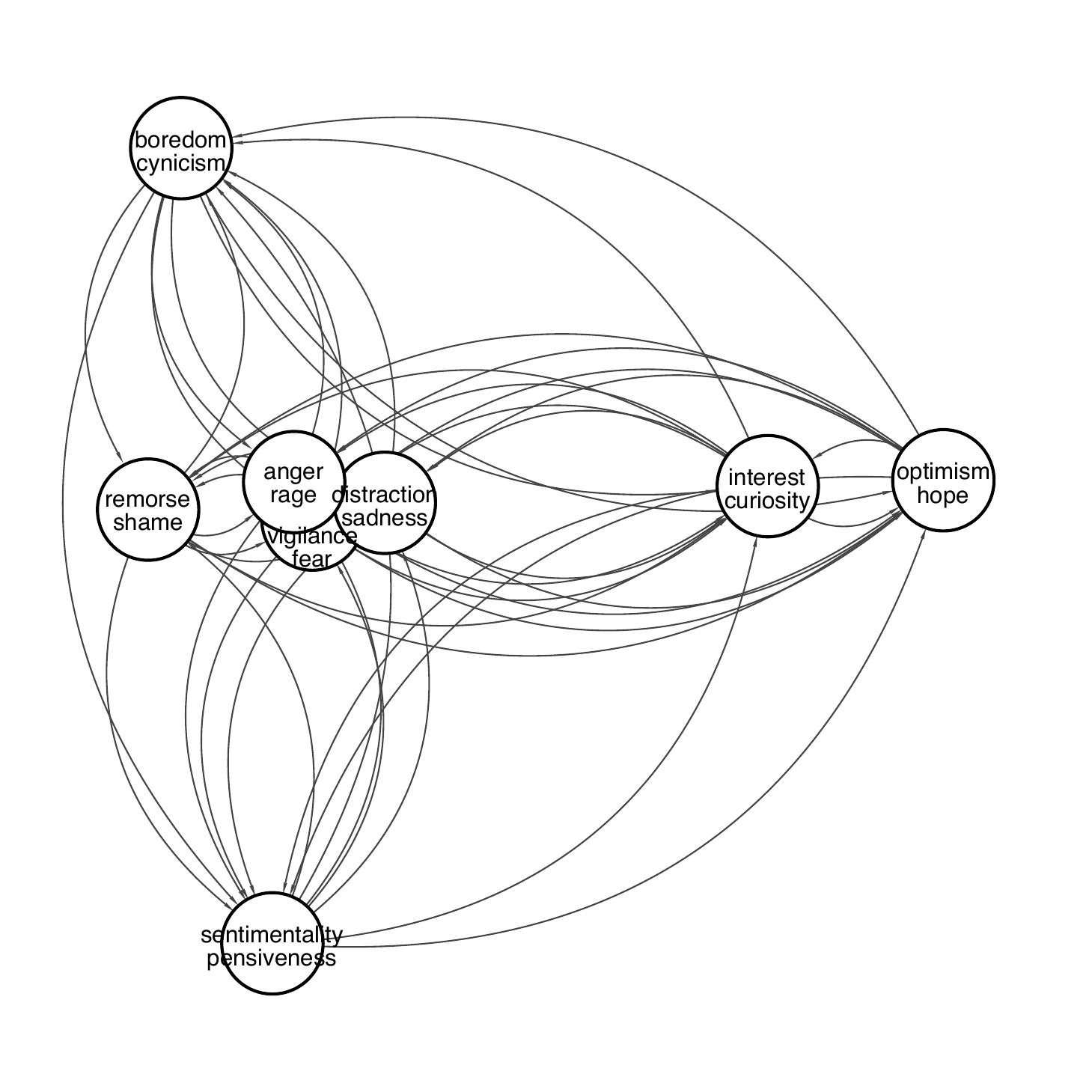}
    \caption{The network of community based on the association}
    \label{fig:asc_communities_network}
\end{figure}

We created the networks whose nodes are a community and whose weight of edges are the value of $\Omega$. 
Since the value of $\Omega$ were small, we multiplied the whole of the value by 10,000. 
We plotted the networks with MDS. 
Then, as the label of each community, we displayed up to the top two emotions with the highest $p(i|k)$ for each community in the MDMC calculation process.
A attribution probability $p(i|k)$ means the membership probability of node $i$ under a condition of community $k$.
We showed the figures to Fig. \ref{fig:sim_communities_network} and \ref{fig:asc_communities_network}.

\subsection{Decomposing the networks with several resolutions}
\label{sec:decomposing-some-resolutions}
MDMC can adjust the number of communities by changing the resolution parameter $\alpha$. 
We performed community extraction at multiple resolution by varying $\alpha$.
First, we showed the results of decomposition of the networks into three communities in Table \ref{tab:sim_comm_3} and \ref{tab:asc_comm_3}.
\begin{table}[htbp]
    \caption{The state of the communities in similarity network (3 communities)}
    \begin{tabular}{ll}
        \hline\noalign{\smallskip}
         k & member \\ 
         \hline\noalign{\smallskip}
         \multirow{4}{*}{1} & joy trust surprise anticipation ecstasy serenity \\
         & admiration acceptance amazement interest optimism \\
         & hope love delight curiosity awe aggressiveness \\
         & pride dominance \\ 
         \hline\noalign{\smallskip}
         \multirow{4}{*}{2} & fear sadness terror apprehension distraction grief \\
         & pensiveness boredom vigilance anxiety guilt \\
         & submission sentimentality despair shame remorse \\
         & pessimism morbidness \\ 
         \hline\noalign{\smallskip}
         \multirow{2}{*}{3} & disgust anger loathing rage annoyance disappointment \\
         & unbelief outrage envy contempt cynicism \\ 
         \hline\noalign{\smallskip}
    \end{tabular}
    \label{tab:sim_comm_3}
\end{table}
\begin{table}[htbp]
    \centering
    \caption{The state of the communities in association network (3 communities)}
    \begin{tabular}{ll}
        \hline\noalign{\smallskip}
         k & member \\ 
         \hline\noalign{\smallskip}
         \multirow{4}{*}{1} & joy trust surprise anticipation ecstasy serenity \\
         & admiration acceptance amazement interest optimism \\
         & hope love delight curiosity awe aggressiveness \\
         & pride dominance \\ 
         \hline\noalign{\smallskip}
         \multirow{3}{*}{2} & fear sadness terror apprehension distraction grief \\
         & pensiveness vigilance anxiety guilt submission \\
         & sentimentality despair pessimism \\ 
         \hline\noalign{\smallskip}
         \multirow{3}{*}{3} & disgust anger loathing boredom rage annoyance \\
         & shame disappointment unbelief outrage remorse envy \\
         & contempt cynicism morbidness \\ 
         \hline\noalign{\smallskip}
    \end{tabular}
    \label{tab:asc_comm_3}
\end{table}
Both networks were decomposed to ``joy, trust" community, ``fear, sadness" community and ``disgust, anger" community when the number of communities was three. 
We showed the state of the communities when the number of communities was two in Tab.\ref{tab:comm_2}.
\begin{table}[htbp]
    \centering
    \caption{The state of the communities in the networks of similarity and association (2 communities)}
    \begin{tabular}{ll}
        \hline\noalign{\smallskip}
         k & member \\ 
         \hline\noalign{\smallskip}
         \multirow{4}{*}{1} & joy trust surprise anticipation ecstasy serenity \\
         & admiration acceptance amazement interest optimism \\
         & hope love delight curiosity awe \\
         & aggressiveness pride dominance \\ 
         \hline\noalign{\smallskip}
         \multirow{6}{*}{2} & fear sadness disgust anger terror apprehension \\
         & distract grief pensiveness loathing boredom rage \\
         & annoyance vigilance anxiety guilt submission \\
         & sentimentality despair shame disappointment unbelief \\
         & outrage remorse envy pessimism contempt \\
         & cynicism morbidness \\ 
         \hline\noalign{\smallskip}
    \end{tabular}
    \label{tab:comm_2}
\end{table}
When we decomposed the two networks, the similarity network and the association network, the results were the same.

\section{Discussion}
\label{sec:discussion}
As we showed some in Section \ref{sec:differences-sim-asc}, similarity and association had differences in their nature. 
In terms of both locality and globality, the result for association was higher than that of similarity.
One interpretation would be that people perceive association more readily than similarity between two emotion words. 
In Section \ref{sec:decomposing-some-resolutions}, when we extracted communities from the similarity network and the association network in low resolution, the results were consistent between the two.
It means that similarity and association have different properties but we cannot detect them when we observe the networks at low resolution.

The results described in Section \ref{sec:decompose-network} suggests that both the structures of the similarity network and the association network were for the most part similar to the structure of the wheel of emotion. 
From Fig. \ref{fig:sim_network_8} and \ref{fig:asc_network_8}, we can see that emotions like ``anger," ``fear," and ``sadness" and their derived emotions are positioned close to each other in both networks, in a similar manner with the wheel of emotion. 
But as to ``joy," ``trust," ``anticipation," ``disgust," and ``surprise" and their derived emotions, we cannot say that they are located nearby to each other. 
Although in the wheel ``surprise" is located closer to ``fear" than to ``joy," in the networks we made, ``surprise" is located more closer to ``joy" than to ``fear". 
From Table \ref{tab:comm_2}, we can see that when we extracted communities, in low resolution, into two communities, the pairs of a primary emotion and a derived emotion in the wheel, namely ``anticipation" and ``vigilance" as well as ``surprise" and ``distraction", were classified into separate communities. We think that it also suggests a difference between the wheel of emotion and the two networks we obtained.
These results suggest that the limitation of representing the structure of emotions as a circumplex model.
We think that strong and weak derived emotions are not just enhanced and weakened expressions of primary emotions, but also be some off from primary emotions.

We showed the relationship between communities by MDS in Section \ref{sec:relationship-communities}. 
In Fig. \ref{fig:sim_communities_network} and \ref{fig:asc_communities_network}, we can see that the community of ``anger," ``sadness" and ``fear" and the community of ``joy" and ``optimism" are located on the right and left of the figure, and they have considerable distance from each other. 
We suppose the reason of this that the network are, at low resolution, split into two communities, one positive and the other negative, such as ``joy" and ``sadness." 
We can see that the small communities such as ``boredom," ``cynicism" and ``sentimentality," which is located above and below of the graph, were closer to ``sadness" then to ``joy."
The results of this visualizing and the analysis of section \ref{sec:decomposing-some-resolutions} suggest that negative emotions have more complicated structure than positive emotions.
We consider that the structure of positive emotions and that of negative emotions are not symmetric.
This results also indicates the limitation of expressing the structure of emotions in a circular pattern.

Lastly, we discuss the possibility that people in \newline Japanese crowdsourcing platform (crowd workers) have some biases.
Nakamura and Majima have had Japanese crowd workers and Japanese students do some cognitive tasks to measure influences of samples and environments \cite{nakamura2019}.
They argued from the results that an experiment that employs Japanese crowd workers online is a valid method for psychological experiments.
In a flanker task they assigned, they could not identify significant difference in reaction time or error rate between crowd workers and students.
On the other hand, Majima observed differences in reaction time and error rate between the two groups in a flanker task he assigned \cite{majima2017}.
The number of attempts of a flanker task by Nakamura and Majima was 40, but that of by Majima was 100.
Given those results, Nakamura and Majima considered that participants tended to cut corners when they worked on monotonous tasks for a long time, and that this tendency could cause a greater differences depending on samples or environments.
We haven't assigned all of the ordered pairs of emotion words to each of participants.
We divided the questions into small sets and assigned them to separate participants.
With these precautions, we think that we have dealt with the concerns about conducting cognitive experiments on a crowdsourcing platform properly.

\section{Conclusion}
\label{sec:conclusion}
In this study, we conducted the experiments that asked participants about similarity and association of the ordered pairs of emotion words, and from the results we made the semantic networks of the emotion words. 
We then analyzed the structure of these networks using community decomposition to compare them to the structure of Plutchik's wheel of emotion. 
The results showed that the structures of the networks and the emotion model were largely similar, with local differences. 
For example, in the wheel of emotion, ``surprise" is an emotion located on the side of ``sadness," whereas in the networks we made it is located on the side of ``joy." 
When we decomposed the networks at low resolution, there was places where the primary emotions and their derived emotions belonged to separate communities. 
The results showed the limitation of representing the emotions in a circular model.
In Plutchik's wheel of emotion, the positive emotions and negative emotions are in opposite positions.
However, our results suggested that such a symmetry is not completely valid and the negative emotions have more complex structure.
We think that focusing on the relationships between emotions is crucial for expressing human emotions in a more natural way.
In this regard, we think that representing structures of emotions as a network is an effective approach.
In order to implement a more natural and accessible HAI technologies, internal representation of emotions that takes into account the interactions between them is an essential part.
We implemented the experiments in Japanese in this research.
It will be meaningful to conduct the experiments in English and other languages to take into account the linguistic and cultural differences.

\bibliographystyle{plain}
\bibliography{ref}

\appendix
\newpage
\section{The Japanese translations of emotion words}

Here are the Japanese translations of the emotion words used in the experiments.
They are described in English (in Japanese) .

\begin{table}[htbp]
    \centering
    \caption{48 emotion words proposed by Plutchik with Japanese translation.}
    \begin{tabular}{ll}
        \hline\noalign{\smallskip}
        \multirow{5}{*}{primary emotions} & joy (yorokobi), trust (shinrai) \\
        & fear (osore), surprise (odoroki) \\
        & sadness (kanashimi) \\
        & disgust (kenno), anger (ikari) \\
        & anticipation (kitai) \\
        \noalign{\smallskip}\hline
        \multirow{7}{*}{strong derived emotions} & ecstasy (koukotsu) \\
        & admiration (kantan) \\
        & terror (kyouhu) \\ 
        & amazement (kyoutan) \\
        & grief (hituu) \\
        & loathing (zouo), rage (gekido) \\
        & vigilance (keikai) \\
        \noalign{\smallskip}\hline
        \multirow{8}{*}{weak derived emotions} & serenity (heion) \\
        & acceptance (younin) \\
        & apprehension (shinpai) \\
        & distraction (douyou) \\
        & pensiveness (urei) \\
        & boredom (taikutsu) \\
        & annoyance (iradachi) \\
        & interest (kyoumi) \\
        \noalign{\smallskip}\hline
        \multirow{16}{*}{secondary emotions} & optimism (rakkan) \\
        & hope (kibou), anxiety (huan) \\
        & love (ai), guilt (zaiakukan) \\
        & delight (kanki) \\
        & submission (hukuju) \\
        & curiosity (koukishin) \\
        & sentimentality (kansyou), awe (ikei) \\
        & despair (zetsubou), shame (haji) \\
        & disappointment (shitsubou) \\
        & unbelief (hushin), outrage (hungai) \\
        & remorse (jiseki), envy (syokubou) \\
        & pessimism (hikan) \\
        & contempt (keibetsu) \\
        & cynicism (reisyou), morbidness (innutsu) \\
        & aggressiveness (sekkyokusei) \\
        & pride (hokori), dominance (yuui) \\
        \noalign{\smallskip}\hline
    \end{tabular}
    \label{tab:emotions_japanese}
\end{table}

\end{document}